\pdfoutput=1

\documentclass[11pt]{article}

\usepackage[table]{xcolor}
\usepackage{acl}
\usepackage{times}
\usepackage{latexsym}
\usepackage{float}
\usepackage[T1]{fontenc}
\usepackage[utf8]{inputenc}
\usepackage{microtype}
\usepackage{graphicx}
\usepackage{arydshln}
\usepackage{longtable}
\usepackage{rotating}
\usepackage{tablefootnote}
\usepackage{multirow}
\usepackage{multicol}
\usepackage{booktabs}
\usepackage{makecell}
\usepackage{arydshln}
\usepackage{xspace}
\usepackage{color,soul}
\usepackage{tabularx}
\usepackage{inconsolata}
\usepackage{amssymb}
\usepackage{graphicx}
\usepackage{calc}
\usepackage{enumitem}
\usepackage{caption}
\usepackage{subcaption}
\usepackage{amsmath}
\usepackage{cleveref}

\newcommand{\ie}{i.e.~}
\newcommand{\eg}{e.g.~}

\newcommand{\mconer}{\mbox{\textsc{MultiCoNER 2}}\xspace} 
\newcommand{\mconerdata}{\mbox{\textsc{MultiCoNER v2}}\xspace} 
\newcommand{\numclasses}{\mbox{33}\xspace} 
         
\newcommand{\facility}{\mbox{\textsc{Facility}}\xspace} 
\newcommand{\scientist}{\mbox{\textsc{Scientist}}\xspace} 
\newcommand{\otherper}{\mbox{\textsc{OtherPER}}\xspace} 
\newcommand{\artist}{\mbox{\textsc{Artist}}\xspace} 
\newcommand{\writtenwork}{\mbox{\textsc{WrittenWork}}\xspace} 
\newcommand{\athlete}{\mbox{\textsc{Athlete}}\xspace} 
\newcommand{\politician}{\mbox{\textsc{Politician}}\xspace} 
\newcommand{\artwork}{\mbox{\textsc{ArtWork}}\xspace} 
\newcommand{\visualwork}{\mbox{\textsc{VisualWork}}\xspace} 
\newcommand{\musicalwork}{\mbox{\textsc{MusicalWork}}\xspace} 
\newcommand{\org}{\mbox{\textsc{ORG}}\xspace} 
\newcommand{\otherprod}{\mbox{\textsc{OtherPROD}}\xspace} 
\newcommand{\otherloc}{\mbox{\textsc{OtherLOC}}\xspace} 
\newcommand{\station}{\mbox{\textsc{Station}}\xspace} 
\newcommand{\vehicle}{\mbox{\textsc{Vehicle}}\xspace} 
\newcommand{\cleric}{\mbox{\textsc{Cleric}}\xspace} 
\newcommand{\musicalgrp}{\mbox{\textsc{MusicalGRP}}\xspace} 
\newcommand{\software}{\mbox{\textsc{Software}}\xspace} 
\newcommand{\humansettlement}{\mbox{\textsc{HumanSettlement}}\xspace} 
\newcommand{\sportsgrp}{\mbox{\textsc{SportsGRP}}\xspace} 
\newcommand{\sportsmanager}{\mbox{\textsc{SportsManager}}\xspace} 
 
\newcommand{\food}{\mbox{\textsc{Food}}\xspace} 
\newcommand{\disease}{\mbox{\textsc{Disease}}\xspace} 
\newcommand{\publiccorp}{\mbox{\textsc{PublicCORP}}\xspace} 
\newcommand{\clothing}{\mbox{\textsc{Clothing}}\xspace} 
\newcommand{\carmanufacturer}{\mbox{\textsc{CarManufacturer}}\xspace} 
\newcommand{\drink}{\mbox{\textsc{Drink}}\xspace} 
\newcommand{\medication}{\mbox{\textsc{Medication/Vaccine}}\xspace} 
\newcommand{\medicalprocedure}{\mbox{\textsc{MedicalProcedure}}\xspace} 
\newcommand{\anatomicalstructure}{\mbox{\textsc{AnatomicalStructure}}\xspace} 
 
\newcommand{\privatecorp}{\mbox{\textsc{PrivateCORP}}\xspace} 
\newcommand{\symptom}{\mbox{\textsc{Symptom}}\xspace} 
\newcommand{\aerospacemanufacturer}{\mbox{\textsc{AerospaceManufacturer}}\xspace}

\newcommand{\exampleq}[1]{``\texttt{{#1}}''}
\newcommand{\langid}[1]{\texttt{#1}}

\usepackage{tasks}
\usepackage[many]{tcolorbox}

\newcommand\team[1]{\mbox{\fontfamily{qpl}\selectfont \textbf{#1}}}
\newcommand\classname[1]{\mbox{\fontfamily{qpl}\selectfont #1}}
\newcommand\lang[1]{\mbox{\fontfamily{qpl}\selectfont #1}}

\title{SemEval-2023 Task 2: Fine-grained Multilingual Named Entity Recognition (MultiCoNER 2)}

\author{Besnik Fetahu~~~~~~Sudipta Kar~~~~~~Zhiyu Chen~~~~~~Oleg Rokhlenko~~~~~~Shervin Malmasi \\
  Amazon.com, Inc.~~~~ Seattle, WA, USA \\
{\{\texttt{besnikf,sudipkar,zhiyuche,olegro,malmasi\}@amazon.com}}}

\begin{document}
\maketitle

\begin{abstract}
\looseness=-5
We present the findings of SemEval-2023 Task 2 on Fine-grained Multilingual Named Entity Recognition (\mconer).\footnote{\url{https://multiconer.github.io}}
Divided into 13 tracks, the task focused on methods to identify complex fine-grained named entities (like \writtenwork, \vehicle, \musicalgrp) across 12 languages, in both monolingual and multilingual scenarios, as well as noisy settings.
The task used the \mconerdata dataset, composed of  2.2 million instances in Bangla, Chinese, English, Farsi, French, German, Hindi, Italian., Portuguese, Spanish, Swedish, and Ukrainian.

\mconer was one of the most popu\-lar tasks of SemEval-2023. It attracted 842 submissions from 47 teams,
and 34~teams~sub\-mitted system papers.
Results showed that complex entity types such as media titles and product names were the most challenging. Methods fusing external knowledge into transformer models achieved the best performance, and the largest gains were on the \classname{Creative Work} and \classname{Group} classes, which are still challenging even with external knowledge. Some fine-grained classes proved to be more challenging than others, such as \scientist, \artwork, and \privatecorp.
We also observed that noisy data has a significant impact on model performance, with an average drop of 10\% on the noisy subset.
The task highlights the need for future research on improving NER robustness on noisy data containing complex entities.

\end{abstract}

\section{Introduction}\label{sec:introduction}

Complex Named Entities (NE), like the titles of creative works, are not simple nouns and pose challenges for NER systems \citep{ashwini2014targetable}. 
They can take the form of any linguistic constituent, like an imperative clause (\exampleq{Dial M for Murder}), and do not look like traditional NEs (Persons, locations, etc.).
This syntactic ambiguity makes it challenging to recognize them based on context.

We organized the Multilingual Complex NER (MultiCoNER) task \cite{multiconer-report} at SemEval-2022 to address these challenges in 11 languages, receiving a positive community response with 34 system papers.
Results confirmed the challenges of processing complex and long-tail NEs: even the largest pretrained Transformers did not achieve top performance without external knowledge.
The top systems infused transformers with knowledge bases and gazetteers.
However, such solutions are brittle against out of knowledge-base entities and noisy scenarios (e.g. spelling mistakes and typos). For entities with fine-grained classes, apart from the entity surface form, the context is critical in determining the correct class.

\mconer expanded on these challenges by adding fine-grained NER classes, and the inclusion of noisy input. Fine-grained NER requires models to distinguish between sub-types of entities that differ only at the fine-grained level, e.g. \scientist vs. \athlete. In these cases, it is crucial for models to capture the entity's context.
In terms of noise, we assessed how small perturbations in the entity surface form and its context can impact performance. Noisy scenarios are quite common in many applications such as Web search and social media.
These challenges are described in \Cref{tab:ner_challenges}, and our tasks defined below.

\begin{table*}[!ht]
    \centering
    \resizebox{0.85\linewidth}{!}{%
    \begin{tabular}{>{\raggedright}p{3.3cm}|p{15cm}} \hline
      Challenge & Description  \\
      
       \hline %
        \textbf{Fine-grained Entities} &
        The entity type can be different based on the context. For example, a creative work entity \exampleq{Harry Potter and the Sorcerer's Stone} could be s a book or a film, depending on the context.
        \\
        
        \hline %
        \textbf{Noisy NER} \newline & 
        Gazetteer based models would not work for typos (e.g., \exampleq{sony xperia} $\rightarrow$ \exampleq{somy xpria}) or spelling errors (e.g., \exampleq{ford cargo} $\rightarrow$ \exampleq{f0rd cargo}) in entities, degrading significantly their performance. %
        \\
      \hline
      \textbf{Ambiguous Entities and Contexts} & 
          Some NEs are ambiguous: they are not always entities, \eg \exampleq{Inside Out}, \exampleq{Among Us}, and \exampleq{Bonanza} may refer to NEs (a movie, video game, and TV show) in some contexts, but not in others. Such NEs often resemble regular syntactic constituents.
        \\
        \hline
        \textbf{Surface Features}  & %
        Capitalization/punctuation features are large drivers of success in NER \citep{mayhew-etal-2019-ner}, but short inputs (ASR, queries) often lack such surface features. An \underline{uncased} evaluation is needed to assess model performance.
        \\
        \hline

    \end{tabular}}
    \vspace{-0.3cm}
    \caption{\small Challenges addressed by \mconer. 
    }
    \label{tab:ner_challenges}
\end{table*}

\begin{enumerate}[leftmargin=*]
\itemsep0em
    \item \textbf{Monolingual}: NER systems are evaluated on monolingual setting, e.g. models are trained and tested on the same language (12 tracks in total).
    \item \textbf{Multilingual}: NER systems are tested on a multilingual test set, composed from all languages in the monolingual track.
\end{enumerate}

We released the \mconerdata dataset \cite{multiconer2-data} to address the aforementioned challenges.
\mconerdata includes data from Wikipedia which has been filtered to identify difficult low-context sentences, and further post-processed.
The data covers 12 languages, which are used to define the 12 monolingual subsets of the task. Additionally, the dataset has a multilingual subset which has mixed data from all the languages.

\mconer received 842 submissions from 47 teams, and 34 teams submitted system description papers.
Results showed that usage of external data and ensemble strategies played a key role in the strong performance. %
External knowledge brought large improvements on classes containing names of creative works and groups, allowing those systems to achieve the best overall results.

Regarding noisy data, all systems show significant performance drop on the noisy subset, which included simulated typographic errors. Small perturbations to entities had a more negative effect than those to the context tokens surrounding entities. This suggests that current systems may not be robust enough to handle real-world noisy data, and that further research is needed to improve their performance in such scenarios. Finally, NER systems seem to be most robust to noise for \classname{PER}, while most susceptible to noise for \classname{GRP}.

In terms of fine-grained named entity types, we observed that performance was lower than the coarse types due to failure to correctly disambiguate sub-classes such as \athlete vs. \sportsmanager. Some of the most challenging fine-grained classes include \privatecorp, \scientist and \artwork.

\section{\mconerdata Dataset}
\label{sec:dataset}

\begin{figure}
    \centering
    \includegraphics[width=1.0\columnwidth,trim=0 5 60 5, clip]{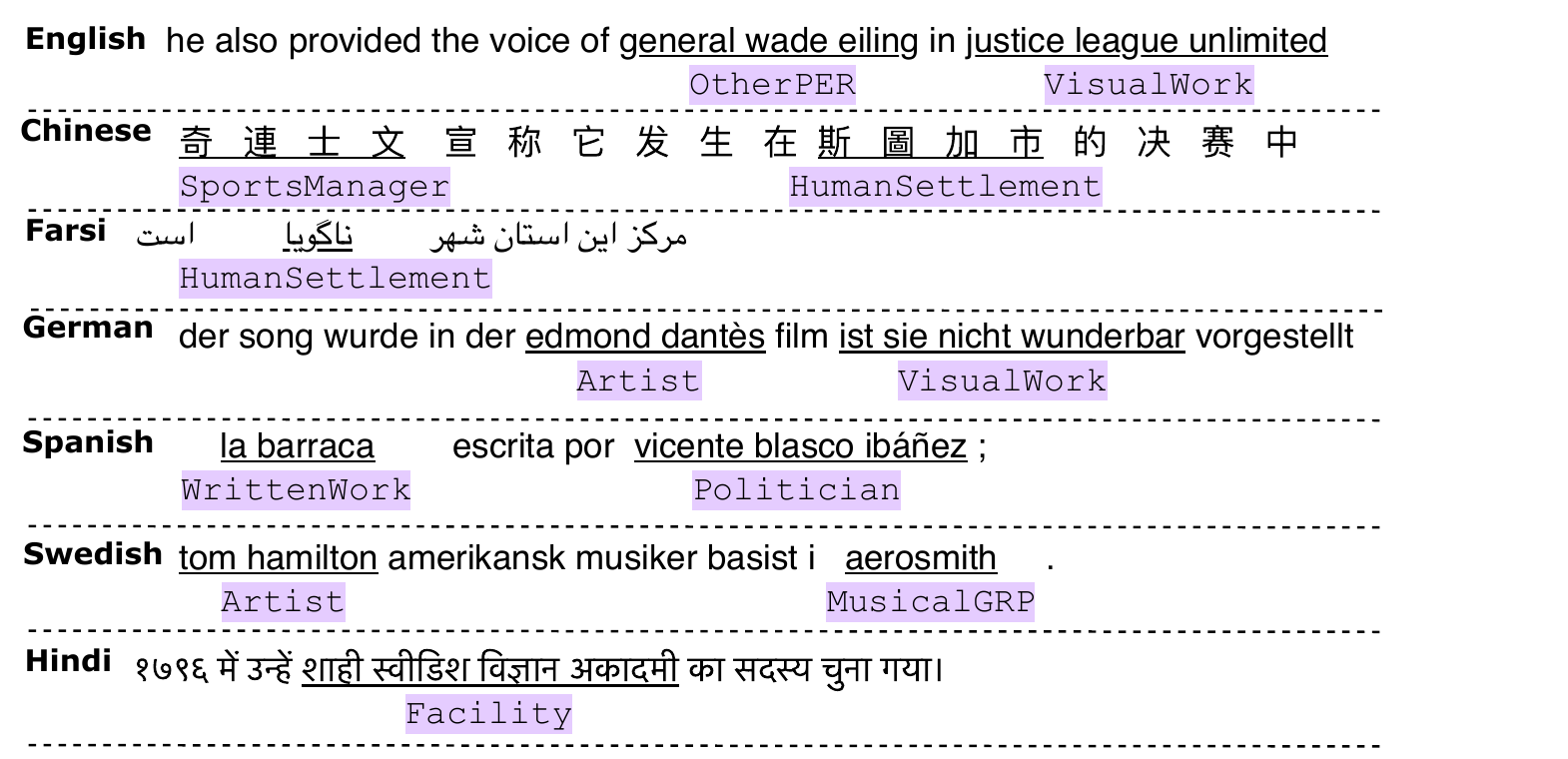}
    \vspace{-20pt}
    \caption{\small Examples sentences from \mconerdata.}
    \label{fig:multiconer2_examples}
\end{figure}

The \mconerdata dataset was designed to address the NER challenges described in \S\ref{sec:introduction}. The data comes from the wiki domain and includes 12 languages, plus a multilingual subset.
Some examples from our data can be seen in Figure~\ref{fig:multiconer2_examples}.
For a detailed description of the \mconerdata data, we refer the reader to the dataset paper~\cite{multiconer2-data}.
The dataset is publicly available.\footnote{\url{https://registry.opendata.aws/multiconer}}

\subsection{Languages and Subsets}
\mconerdata covers 12 languages:\vspace{-10pt}

\begin{table}[H]
    \centering
    \small{
    \begin{tabular}{l@{\hskip .25in}l l }
        \lang{Bangla} (\langid{BN)} & \lang{Chinese} (\langid{ZH)} & \lang{English} (\langid{EN}) \\
         \lang{Farsi} (\langid{FA}) & \lang{French} (\langid{FR}) & \lang{German} (\langid{DE})\\
          \lang{Hindi} (\langid{HI}) & \lang{Italian} (\langid{IT)} & \lang{Portuguese} (\langid{PT})\\
        \lang{Spanish} (\langid{ES})  & \lang{Swedish} (\langid{SV}) & \lang{Ukrainian} (\langid{UK}) \\

    \end{tabular}}
    \label{tab:languages}
\end{table}\vspace{-10pt}

These languages were chosen to include a diverse typology of languages and writing systems, and range from well-resourced (\langid{EN}) to low-resourced ones (\langid{FA}).
\mconerdata contains 13 different subsets: 12 monolingual, and a multilingual subset (denoted as \langid{MULTI}).

\paragraph{Monolingual Subsets}
Each of the 12 languages has its own subset. %

\paragraph{Multilingual Subset}
This contains randomly sampled data from all the languages mixed into a single subset. This subset is designed for evaluating multilingual models, and should ideally be used under the assumption that the language for each sentence is unknown. From the test set of each language, we randomly selected at most 35,000 samples resulting in a total of 358,668 instances.

\subsection{Dataset Creation}
\label{sec:creation}

In this section, we provide a brief overview of the dataset construction process. Additional details are available in \citet{multiconer2-data}.

\mconerdata was extracted following the same strategy as \citet{DBLP:conf/coling/MalmasiFFKR22}, where sentences from the different languages are extracted from localized versions of Wikipedia. We select low-context sentences and the \emph{interlinked} entities are resolved to the \emph{entity types} using Wikidata as a reference, according to the NER class taxonomy shown in Table~\ref{tab:mconer_taxonomy}. Furthermore, to prevent models from leveraging surface form features, we lowercase the words and remove punctuation. These steps result in more challenging sentences that are more representative of real-world data.

\subsection{Fine-grained NER Taxonomy}
\label{sec:taxonomy}

\mconer builds on top of the WNUT 2017 \citep{wnut2017} taxonomy entity types, and adds an additional layer of fine-grained types. We also drop the \classname{Corporation} class, as it overlaps with the \classname{Group} class. Furthermore, we introduce a new coarse grained class called \classname{Medical}, which captures entities from the medical domain (e.g. \disease, \anatomicalstructure, etc.).
Table~\ref{tab:mconer_taxonomy} shows the \numclasses fine-grained classes, grouped across 6 coarse types.

\begin{table*}[hbt!]
    \centering
    \resizebox{0.9\textwidth}{!}{
    \begin{tabular}{l l l l l l}
    \toprule
    \classname{PER} (\classname{Person}) &     \classname{LOC}  (\classname{Location})&     \classname{GRP}  (\classname{Group}) &     \classname{PROD}  (\classname{Product}) &     \classname{CW}  (\classname{Creative Work}) &     \classname{MED}  (\classname{Medical})\\
    \midrule
    \artist     & \facility & {\small \aerospacemanufacturer} & \clothing & \artwork & \anatomicalstructure\\
    \athlete     &  \humansettlement & \carmanufacturer & \drink & \musicalwork & \disease\\
    \cleric     &  \station &  \musicalgrp & \food & \software & \medicalprocedure\\
    \politician     &  \otherloc &  \org & \vehicle & \visualwork & \medication\\
    \scientist     &  & \privatecorp & \otherprod & \writtenwork & \symptom\\
    \sportsmanager     &  & \publiccorp  & & \\
    \otherper     &  & \sportsgrp\\
    \bottomrule
    \end{tabular}}\vspace{-7pt}
    \caption{\small{\mconerdata NER taxonomy, consisting of \numclasses fine-grained classes, grouped across 6 coarse grained types.}}
    \label{tab:mconer_taxonomy}
\end{table*}

The fine-grained taxonomy allows us to capture a wide array of entities, including complex entity structures, such as \classname{CW},
or entities that are ambiguous without their context, e.g. \scientist vs. \athlete as part of the \classname{PER} coarse grained type.

\begin{table*}[t]
\centering
\resizebox{0.85\textwidth}{!}{
    
    \begin{tabular}{l l r r r r r r r r r r r r r}
    \toprule
    \textbf{Class} & \textbf{Split} & \textbf{\langid{EN}} & \textbf{\langid{DE}} & \textbf{\langid{FA}} & \textbf{\langid{FR}} & \textbf{\langid{ES}} & \textbf{\langid{UK}} & \textbf{\langid{SV}} & \textbf{\langid{HI}} &\textbf{ \langid{BN}} & \textbf{\langid{ZH}} & \textbf{\langid{IT}} & \textbf{\langid{PT}} & \textbf{\langid{Multi}}\\
    \midrule
\multirow{3}{*}{PER} & train & 9,294 & 5,508 & 8,006 & 9,295 & 8,360 & 6,441 & 7,695 & 3,609 & 3,778 & 4,862 & 10,387 & 8,241 & 85,476 \\
 & dev & 481 & 280 & 413 & 483 & 442 & 341 & 445 & 174 & 194 & 239 & 548 & 447 & 4,487 \\
 & test &137,681 & 11,299 & 115,868 & 141,401 & 125,379 & 96,864 & 111,157 & 5,736 & 6,935 & 9,095 & 160,598 & 120,413 & 180,080 \\
\midrule
\multirow{3}{*}{CW} & train & 4,084 & 2,466 & 3,661 & 5,438 & 3,606 & 2,907 & 3,714 & 1,646 & 1,981 & 2,264 & 5,048 & 3,839 & 40,654 \\
 & dev & 215 & 127 & 184 & 268 & 183 & 146 & 200 & 90 & 103 & 112 & 267 & 206 & 2,101 \\
 & test & 62,126 & 4,777 & 53,034 & 84,952 & 55,459 & 43,291 & 54,806 & 2,804 & 3,640 & 4,369 & 79,873 & 58,245 & 87,030 \\
\midrule
\multirow{3}{*}{GRP} & train & 4,224 & 2,815 & 3,209 & 3,745 & 3,632 & 3,204 & 3,459 & 2,273 & 2,227 & 2,696 & 3,416 & 3,788 & 38,688 \\
 & dev & 218 & 177 & 180 & 195 & 195 & 151 & 194 & 143 & 122 & 145 & 173 & 200 & 2,093 \\
 & test &60,026 & 4,418 & 38,807 & 52,987 & 50,259 & 39,709 & 46,929 & 3,897 & 3,651 & 4,715 & 46,271 & 48,994 & 73,226 \\
\midrule
\multirow{3}{*}{LOC} & train & 4,353 & 2,269 & 5,086 & 4,723 & 4,651 & 5,458 & 7,176 & 2,487 & 2,457 & 2,470 & 4,446 & 4,794 & 50,370 \\
 & dev & 197 & 117 & 267 & 242 & 230 & 294 & 370 & 133 & 127 & 129 & 248 & 250 & 2,604 \\
 & test &67,901 & 5,306 & 70,907 & 73,373 & 72,996 & 84,643 & 111,879 & 7,172 & 7,375 & 6,170 & 68,564 & 70,923 & 117,257 \\
\midrule
\multirow{3}{*}{PROD} & train & 1,935 & 1,571 & 2,049 & 1,946 & 1,989 & 2,258 & 1,989 & 1,420 & 1,384 & 1,529 & 1,770 & 1,927 & 21,767 \\
 & dev & 109 & 78 & 107 & 100 & 100 & 117 & 112 & 74 & 67 & 73 & 86 & 101 & 1,124 \\
 & test &27,580 & 1,643 & 18,212 & 28,274 & 28,469 & 30,071 & 22,686 & 1,611 & 1,493 & 1,869 & 22,887 & 21,115 & 35,545 \\
\midrule
\multirow{3}{*}{MED} & train & 1,559 & 1,322 & 1,651 & 1,230 & 1,669 & 1,688 & 1,381 & 1,435 & 1,396 & 1,407 & 1,376 & 1,850 & 17,964 \\
 & dev & 76 & 62 & 85 & 64 & 81 & 86 & 70 & 70 & 63 & 75 & 76 & 88 & 896 \\
 & test &22,491 & 1,434 & 15,287 & 17,208 & 23,812 & 20,796 & 13,702 & 1,979 & 1,919 & 1,781 & 19,029 & 21,062 & 29,553 \\
\midrule
\multirow{3}{*}{\textbf{Total}} & train & 16,778 & 9,785 & 16,321 & 16,548 & 16,453 & 16,429 & 16,363 & 9,632 & 9,708 & 9,759 & 16,579 & 16,469 & 170,824 \\
 & dev & 871 & 512 & 855 & 857 & 854 & 851 & 856 & 514 & 507 & 506 & 858 & 854 & 8,895 \\
 & test &249,980 & 20,145 & 219,168 & 249,786 & 246,900 & 238,296 & 231,190 & 18,399 & 19,859 & 20,265 & 247,881 & 229,490 & 358,668 \\

    \bottomrule
    \end{tabular}}\vspace{-7pt}
    \caption{\small{\mconer dataset statistics for the different languages for the Train/Dev/Test splits. For each NER class we show the total number of entity instances per class on the different data splits. The bottom three rows show the total number of sentences for each language.}}
    \label{tab:data_stats_coarse}
\end{table*}

\subsection{Noisy Subsets}
\label{sec:data_noise}

NER systems are typically trained on carefully curated datasets. However, in real-world scenarios, various errors may arise due to human mistakes. We applied noise only on the test set to simulate environments where NER models are exposed directly to user-generated content.

To evaluate the robustness of NER models, we corrupt 30\% of the test set
with various types of simulated errors in 7 languages (\langid{EN}, \langid{ZH}, \langid{IT}, \langid{ES}, \langid{FR}, \langid{PT}, \langid{SV}). The corruption can impact context tokens and entity tokens.
For Chinese, we applied character-level corruption strategies~\cite{wang2018hybrid} which involve replacing characters with visually or phonologically resembled ones. For other languages, we developed token-level corruption strategies based on common typing mistakes made by humans (e.g., randomly substituting a letter with a neighboring letter on the keyboard), utilizing language specific keyboard layouts.\footnote{We extended the keyboard layouts in this library to include 7 languages: \tiny\url{https://github.com/ranvijaykumar/typo}}

\subsection{Dataset Statistics}
Table \ref{tab:data_stats_coarse} shows the \mconerdata dataset statistics.
For most tracks, we released 16k training and 800 development instances (with the exception of \langid{DE}, \langid{BN}, \langid{HI}, \langid{ZH} due to data scarcity).

The test splits on the other hand are much larger. This is done for two reasons: (1) to assess the generalizability of NER models in identifying unseen and complex fine-grained entity types, where the entity overlap between train and test sets is small, and
and (2) to assess how models handle noise in contextual or entity tokens.
For practical reasons, we cap the number of test instances to be less than 250k per subset for most languages (with the exception of \langid{DE}, \langid{BN}, \langid{HI}, \langid{ZH} which are already small due to data scarcity).

\section{Task Description and Evaluation}
\label{sec:task_description}
The shared task is composed of 12 monolingual and 1 multilingual track.
The multilingual track invited multilingual models capable of identifying entities from monolingual texts from any of the 12 languages.
As described in Section~\ref{sec:data_noise}, 30\% of the test sets of the \langid{EN}, \langid{ZH}, \langid{IT}, \langid{ES}, \langid{FR}, \langid{PT}, and \langid{SV} monolingual tracks are corrupted with simulated noise. We refer the subsets with corruption as \textbf{noisy subsets} and the rest as \textbf{clean subsets}.

For evaluation, we used the macro-averaged F1 scores to evaluate and rank systems. The F1 scores are computed over the fine-grained types using exact matching (\ie the entity boundary and type must exact match the ground truth), and averaged across all types.
We also report the performance on noisy subsets and clean subsets in Appendix \ref{sec:app_detail_results} to study the impact on noise in \S\ref{sec:results}.

\section{Baseline System}\label{sec:baseline}
Similar to the 2022 edition~\cite{multiconer-report}, we train and evaluate a baseline NER system using XLM-RoBERTa (XLM-R)~\cite{conneau-etal-2020-unsupervised}, a multilingual Transformer model. The XLM-R model computes a representation for each token, which is then used to predict the token tag using a CRF classification layer~\cite{sutton2012introduction}.

XLM-R is suited for multilingual scenarios, supporting up to 100 languages. It provides a solid baseline upon which the participants can build. It was trained with a learning rate of $2e-5$ and for 50 epochs, with an early stopping criterion of a non-decreasing validation loss for 5 epochs. 
The code and scripts for the baseline system were provided to the participants to use its functionalities and further extend it with their approaches.\footnote{{\scriptsize\url{https://github.com/amzn/multiconer-baseline}}}

\section{Participating Systems and Results}
\label{sec:participating_systems}

\begin{table*}[!hbtp]
\centering
\resizebox{0.95\textwidth}{!}{%
\begin{tabular}{|lll|lll|lll|lll|}

\hline

\multicolumn{3}{|c|}{\cellcolor{cyan!25}\textbf{English (EN})} & 10 & silp\_nlp & 65.00 & 4 & NLPeople & 70.76 & 10 & BizNER & 67.71 \\ 
1 & DAMO-NLP & 85.53 & 11 & LSJSP & 64.36 & 5 & IXA/Cogcomp & 69.49 & 11 & LLM-RM & 63.29 \\ 
2 & SRC - Beijing & 83.09 & 12 & D2KLab & 62.98 & 6 & USTC-NELSLIP & 68.85 & 12 & D2KLab & 63.29 \\
3 & PAI & 80.00 & 13 & Sartipi-Sedighin & 61.95 & 7 & PAI & 68.46 & 13 & Sartipi-Sedighin & 63.10 \\ 
4 & CAIR-NLP & 79.33 & 14 & SAB & 58.03 & 8 & Sakura & 64.88 & 14 & SAB & 62.30 \\ 
5 & KDDIE & 78.06 &  & \textbf{BASELINE} & 57.19 & 9 & garNER & 62.12 & 15 & LSJSP & 53.13 \\ 
6 & SRCB & 75.62 & 15 & FII\_Better & 52.12 & 10 & Sartipi-Sedighin & 60.02 & 16 & L3i++ & 43.56 \\ 
7 & IXA/Cogcomp & 72.82 & 16 & IXA & 25.96 & 11 & D2KLab & 54.20 & 17 & IXA & 26.13 \\ 
8 & USTC-NELSLIP & 72.15 & \multicolumn{3}{|c|}{\cellcolor{cyan!25}\textbf{Ukranian (UK})} & 12 & Ertim & 53.77 & \multicolumn{3}{|c|}{\cellcolor{cyan!25}\textbf{Bangla (BN})} \\ 
9 & NLPeople & 71.81 & 1 & DAMO-NLP & 89.02 & 13 & SAB & 52.42 & 1 & PAI & 84.39 \\ 
10 & BizNER & 70.44 & 2 & CAIR-NLP & 81.29 &  & \textbf{BASELINE} & 51.56 & 2 & DAMO-NLP & 81.60 \\ 
11 & Sakura & 70.16 & 3 & IXA/Cogcomp & 75.25 & 14 & IXA & 15.87 & 3 & USTC-NELSLIP & 80.59 \\ 
12 & RIGA & 69.30 & 4 & USTC-NELSLIP & 74.37 & \multicolumn{3}{|c|}{\cellcolor{cyan!25}\textbf{German (DE})} & 4 & IXA/Cogcomp & 78.95 \\ 
13 & CodeNLP & 63.51 & 5 & NLPeople & 73.41 & 1 & PAI & 88.09 & 5 & NLPeople & 78.24 \\ 
14 & Sartipi-Sedighin & 63.25 & 6 & Sakura & 72.31 & 2 & DAMO-NLP & 84.97 & 6 & Sakura & 77.20 \\ 
15 & IITD & 63.21 & 7 & PAI & 71.28 & 3 & IXA/Cogcomp & 80.35 & 7 & MLlab4CS & 76.27 \\ 
16 & garNER & 62.73 & 8 & Sartipi-Sedighin & 67.25 & 4 & USTC-NELSLIP & 78.71 & 8 & garNER & 73.39 \\ 
17 & FII\_Better & 61.75 & 9 & garNER & 65.64 & 5 & NLPeople & 77.67 & 9 & silp\_nlp & 73.22 \\ 
18 & D2KLab & 61.29 & 10 & D2KLab & 64.14 & 6 & Sakura & 76.24 & 10 & CAIR-NLP & 69.46 \\ 
19 & silp\_nlp & 60.85 & 11 & silp\_nlp & 63.18 & 7 & CAIR-NLP & 74.71 &  & \textbf{BASELINE} & 68.24 \\ 
20 & Ertim & 59.03 & 12 & SAB & 59.42 & 8 & BizNER & 71.21 & 11 & Sartipi-Sedighin & 64.83 \\ 
21 & MEERQAT-IRIT & 58.70 & 13 & LSJSP & 58.07 &  & \textbf{BASELINE} & 67.21 & 12 & VBD\_NLP & 64.50 \\ 
22 & LSJSP & 57.51 &  & \textbf{BASELINE} & 57.29 & 9 & D2KLab & 67.09 & 13 & BizNER & 64.37 \\ 
23 & RGAT & 56.91 & 14 & IXA & 22.81 & 10 & silp\_nlp & 64.92 & 14 & D2KLab & 61.43 \\ 
24 & CLaC & 55.05 & \multicolumn{3}{|c|}{\cellcolor{cyan!25}\textbf{Portugese (PT})} & 11 & Sartipi-Sedighin & 64.21 & 15 & SAB & 56.01 \\ 
25 & L3i++ & 53.00 & 1 & DAMO-NLP & 85.97 & 12 & garNER & 63.88 & 16 & LSJSP & 55.76 \\ 
 & \textbf{BASELINE} & 52.98 & 2 & PAI & 81.61 & 13 & FII\_Better & 55.86 & 17 & L3i++ & 41.33 \\ 
26 & VBD\_NLP & 52.65 & 3 & CAIR-NLP & 80.16 & 14 & LLM-RM & 55.54 & 18 & IXA & 18.49 \\ 
27 & LLM-RM & 52.08 & 4 & BizNER & 72.97 & 15 & SAB & 55.51 & \multicolumn{3}{|c|}{\cellcolor{cyan!25}\textbf{Italian (IT})} \\ 
28 & Minanto & 51.47 & 5 & IXA/Cogcomp & 72.28 & 16 & L3i++ & 46.55 & 1 & DAMO-NLP & 89.79 \\ 
29 & SAB & 51.41 & 6 & USTC-NELSLIP & 71.26 & 17 & IXA & 16.09 & 2 & PAI & 84.88 \\ 
30 & ShathaTaymaaTeam & 50.02 & 7 & Deep Learning Brasil & 70.97 & \multicolumn{3}{|c|}{\cellcolor{cyan!25}\textbf{Chinese (ZH})} & 3 & CAIR-NLP & 83.78 \\ 
31 & azaad@BND & 47.42 & 8 & NLPeople & 70.16 & 1 & NetEase.AI & 84.05 & 4 & BizNER & 76.48 \\ 
32 & LISAC FSDM-USMBA & 44.00 & 9 & Sakura & 69.98 & 2 & DAMO-NLP & 75.98 & 5 & USTC-NELSLIP & 75.70 \\ 
33 & YNU-HPCC & 28.52 & 10 & garNER & 64.51 & 3 & SRCB & 75.86 & 6 & IXA/Cogcomp & 74.67 \\ 
34 & IXA & 15.39 & 11 & Sartipi-Sedighin & 61.28 & 4 & PAI & 74.87 & 7 & Sakura & 74.19 \\ 
\multicolumn{3}{|c|}{\cellcolor{cyan!25}\textbf{Spanish (ES})} & 12 & silp\_nlp & 61.05 & 5 & Taiji & 72.52 & 8 & NLPeople & 73.71 \\ 
1 & DAMO-NLP & 89.78 & 13 & D2KLab & 60.79 & 6 & USTC-NELSLIP & 66.57 & 9 & garNER & 68.20 \\ 
2 & CAIR-NLP & 83.63 & 14 & MEERQAT-IRIT & 59.87 & 7 & NLPeople & 65.96 & 10 & D2KLab & 64.77 \\ 
3 & USTC-NELSLIP & 74.44 & 15 & LSJSP & 58.23 & 8 & IXA/Cogcomp & 64.86 & 11 & Sartipi-Sedighin & 64.50 \\ 
4 & IXA/Cogcomp & 73.81 & 16 & SAB & 54.12 & 9 & Sakura & 64.61 & 12 & silp\_nlp & 63.11 \\ 
5 & Sakura & 72.85 &  & \textbf{BASELINE} & 53.52 & 10 & garNER & 63.47 &  & \textbf{BASELINE} & 57.71 \\ 
6 & NLPeople & 72.76 & 17 & IXA & 16.97 & 11 & Ertim & 59.45 & 13 & SAB & 57.57 \\ 
7 & PAI & 71.67 & \multicolumn{3}{|c|}{\cellcolor{cyan!25}\textbf{French (FR})} & 12 & Sartipi-Sedighin & 58.70 & 14 & FII\_Better & 56.36 \\ 
8 & BizNER & 71.48 & 1 & DAMO-NLP & 89.59 & 13 & CAIR-NLP & 58.43 & 15 & IXA & 18.41 \\ 
9 & garNER & 63.73 & 2 & PAI & 86.17 &  & \textbf{BASELINE} & 58.03 & \multicolumn{3}{|c|}{\cellcolor{cyan!25}\textbf{Multilingual (MULTI})} \\ 
10 & D2KLab & 63.17 & 3 & CAIR-NLP & 83.08 & 14 & Janko & 57.90 & 1 & DAMO-NLP & 84.48 \\ 
11 & silp\_nlp & 62.90 & 4 & BizNER & 78.01 & 15 & YNUNLP & 56.57 & 2 & CAIR-NLP & 79.16 \\ 
12 & MEERQAT-IRIT & 60.93 & 5 & IXA/Cogcomp & 74.52 & 16 & D2KLab & 54.92 & 3 & NLPeople & 78.38 \\ 
13 & LSJSP & 60.55 & 6 & USTC-NELSLIP & 74.25 & 17 & silp\_nlp & 51.65 & 4 & IXA/Cogcomp & 78.17 \\ 
14 & Sartipi-Sedighin & 58.41 & 7 & Sakura & 72.86 & 18 & SAB & 44.12 & 5 & PAI & 77.00 \\ 
15 & LLM-RM & 54.81 & 8 & NLPeople & 72.85 & 19 & NCUEE-NLP & 44.09 & 6 & USTC-NELSLIP & 75.62 \\ 
16 & FII\_Better & 54.51 & 9 & Ertim & 66.30 & 20 & L3i++ & 35.34 & 7 & Sakura & 73.82 \\ 
 & \textbf{BASELINE} & 53.43 & 10 & garNER & 65.68 & 21 & YNU-HPCC & 31.66 & 8 & MaChAmp & 73.74 \\ 
17 & SAB & 48.22 & 11 & D2KLab & 64.09 & 22 & IXA & 6.93 & 9 & CodeNLP & 73.22 \\ 
18 & IXA & 16.01 & 12 & silp\_nlp & 62.39 & \multicolumn{3}{|c|}{\cellcolor{cyan!25}\textbf{Hindi (HI})} & 10 & Lumi & 72.15 \\ 
\multicolumn{3}{|c|}{\cellcolor{cyan!25}\textbf{Swedish (SV})} & 13 & MEERQAT-IRIT & 58.90 & 1 & USTC-NELSLIP & 82.14 & 11 & Sartipi-Sedighin & 71.79 \\ 
1 & DAMO-NLP & 89.57 & 14 & LSJSP & 56.83 & 2 & PAI & 80.96 & 12 & garNER & 69.16 \\ 
2 & CAIR-NLP & 82.88 &  & \textbf{BASELINE} & 55.91 & 3 & IXA/Cogcomp & 79.56 & 13 & LEINLP & 64.63 \\ 
3 & IXA/Cogcomp & 76.54 & 15 & SAB & 55.07 & 4 & DAMO-NLP & 78.56 & 14 & D2KLab & 63.83 \\ 
4 & BizNER & 76.12 & 16 & Sartipi-Sedighin & 54.94 & 5 & NLPeople & 78.50 &  & \textbf{BASELINE} & 62.86 \\ 
5 & USTC-NELSLIP & 75.47 & 17 & IXA & 17.40 & 6 & Sakura & 78.37 & 15 & SAB & 59.55 \\ 
6 & NLPeople & 75.08 & \multicolumn{3}{|c|}{\cellcolor{cyan!25}\textbf{Farsi (FA})} & 7 & silp\_nlp & 74.32 & 16 & LSJSP & 51.74 \\ 
7 & Sakura & 73.79 & 1 & DAMO-NLP & 87.93 & 8 & CAIR-NLP & 72.23 & 17 & SibNN & 50.55 \\ 
8 & PAI & 72.38 & 2 & CAIR-NLP & 77.50 & 9 & garNER & 71.23 & 18 & L3i++ & 44.37 \\ 
9 & garNER & 67.63 & 3 & BizNER & 73.49 &  & \textbf{BASELINE} & 71.20 &  & &  \\

\hline
\end{tabular}}
\caption{\small Rankings for all tracks based on Macro F1. The ``SRC - Beijing'' team is ``Samsung Research China - Beijing''. 
}
\label{tab:f1-based-ranking-full}
\end{table*}
We have received submissions from 47 different teams. Table \ref{tab:f1-based-ranking-full} shows the final rankings for all tracks (fine-grained Macro F1).
Among the monolingual tracks, we have observed the highest participation in the \langid{English} track with 34 teams.
Ordered by the number of participating teams, the rest of the monolingual tracks are \langid{Chinese} (22), \langid{German} (17), \langid{Bangla} (18), \langid{Spanish} (18), \langid{Hindi} (17), \langid{French} (17), \langid{Portuguese} (17), \langid{Swedish} (16), \langid{Italian} (15), \langid{Farsi} (14),  and \langid{Ukrainian} (14).
The number of participating teams for the \langid{Multilingual} track is 18.
Detailed performance breakdown for noisy and clean subsets
of \langid{English}, \langid{Spanish}, \langid{French}, \langid{Italian}, \langid{Portuguese}, \langid{Swedish}, and \langid{Chinese} is available in Appendix \ref{sec:app_detail_results}.

\subsection{Top Systems}

\team{DAMO-NLP}~\cite{tan-EtAl:2023:SemEval} ranked $1^{st}$ for most tracks, except being $2^{nd}$ in \langid{BN}, \langid{DE}, \langid{ZH}, and $4^{th}$ in \langid{HI}. They proposed an unified retrieval-augmented system (U-RaNER) for the task. The system uses two different knowledge sources (Wikipedia paragraphs and the Wikidata knowledge graph) to inject additional relevant knowledge to their NER model. Additionally, they explored an infusion approach to provide more extensive contextual knowledge about entities to the model.

\team{PAI}~\cite{ma:2023:SemEval2} ranked $1^{st}$ in \langid{BN}, \langid{DE}, $2^{nd}$ in \langid{FR}, \langid{HI}, \langid{IT}, \langid{PT}, $3^{rd}$ in \langid{EN}, $4^{th}$ in \langid{ZH}, $5^{th}$ in \langid{MULTI}, $7^{th}$ in \langid{ES}, \langid{FA}, \langid{UK}, and $8^{th}$ in \langid{SV}.
They developed a knowledge base using entities and their associated properties like \texttt{``instanceof''}, \texttt{``subclassof''} and \texttt{``occupation''} from Wikidata. For a given sentence, they used a retrieval module to gather different properties of the entities by string matching. They observed benefits on the clean subset through the dictionary fusing approach. The same benefits were not observed on the noisy subset.

\team{USTC-NELSLIP}~\cite{ma-EtAl:2023:SemEval} ranked $1^{st}$ in \langid{HI}, $3^{rd}$ in \langid{BN}, \langid{ES}, $4^{th}$ in \langid{DE}, \langid{UK}, $5^{th}$ in \langid{IT}, \langid{SV}, $6^{th}$ in \langid{FA}, \langid{FR}, \langid{PT}, \langid{ZH}, \langid{MULTI},  and $8^{th}$ in \langid{EN}.
They proposed a two-stage training strategy. In the first stage, the representations of gazetteer network and language model are adapted at sentence and entity level through minimizing the KL divergence between their representations. In the second stage, two networks are trained together on the NER objective. The final predictions are derived from an ensemble of trained models. The results indicate that the gazetteer played a crucial role in accurately identifying complex entities during the NER process, and the implementation of a two-stage training strategy was effective.

\team{NetEase.AI}~\cite{lu-EtAl:2023:SemEval} ranked $1^{st}$ in \langid{ZH}.
Their proposed system consists of multiple modules. First, a BERT model is used to correct any potential errors in the original input sentences.
The NER module takes the corrected text as input and consists of a basic NER module and a gazetteer enhanced NER module. This approach boosted the performance on the entity level noise and gave the system a strong advantage over the other teams (Table \ref{tab:full_rank_zh}). A retrieval system takes the candidate entity as input and retrieves additional context information, which is subsequently used as input to a text classification model to calculate the probability of the entity's type label. A stacking model is trained to output the final prediction based on the features from multiple modules.

\subsection{Other Noteworthy Systems}
\team{CAIR-NLP}~\cite{n-paul-chaudhary:2023:SemEval} ranked $2^{nd}$ in \langid{MULTI}, \langid{ES}, \langid{FA}, \langid{SV}, \langid{UK}, $3^{rd}$ in \langid{FR}, \langid{IT}, \langid{PT}, $4^{th}$ in \langid{EN}, $7^{th}$ in \langid{DE}, $8^{th}$ in \langid{HI}, $10^{th}$ in \langid{BN}, and $13^{th}$ in \langid{ZH}.
They developed a multi-objective joint learning system (MOJLS) that learns an enhanced representation of low-context and fine-grained entities. In their training procedure they minimize for: 1) representation gaps between fine-grained entity types within a coarse grained type, 2) representation gaps between an input sentence and the input augmented with external information for a given entity, 3) negative log-likelihood loss, 4) biaffine layer label prediction loss. Additionally, external context is retrieved via search engines for an input text, as well as  ConceptNet data \cite{DBLP:journals/corr/SpeerCH16} to better represent an entity class with alternative names, aliases, and relation types to other concepts.

\team{SRCB}~\cite{zhang-EtAl:2023:SemEval4} ranked $3^{rd}$ in \langid{ZH} and $6^{th}$ in \langid{EN}. The proposed approach, for an input sentence retrieves external evidence coming from Wikidata and Wikipedia, which is concatenated with the original input using special tokens (e.g. \texttt{``context''}, \texttt{``prompt \& description''}) to allow their models (based on~\cite{li-etal-2020-unified}), to distinguish the different contexts. To retrieve the external context, the authors first detect entity mentions~\cite{su2022global} from the input sentence, then query the corresponding external sources. 

\team{NLPeople}~\cite{elkaref-EtAl:2023:SemEval} ranked $3^{rd}$ in \langid{MULTI}, $4^{th}$ in \langid{FA}, $5^{th}$ in \langid{BN}, \langid{DE}, \langid{HI}, \langid{UK}, $6^{th}$ in \langid{ES}, \langid{SV}, $7^{th}$ in \langid{ZH}, $8^{th}$ in \langid{FR}, \langid{IT}, \langid{PT}, and $8^{th}$ in \langid{EN}.
They developed a two stage approach. First they extract spans that can be entities, while in the second step they classify spans into the most likely entity type. They augmented the training data with external context by adding relevant paragraphs, infoboxes, and titles from Wikipedia. On languages with smaller test sets, the infoboxes showed to obtain better performance than adding relevant paragraphs. 

\team{IXA/Cogcomp}~\cite{garcaferrero-EtAl:2023:SemEval} ranked $3^{rd}$ in \langid{DE}, \langid{HI}, \langid{UK}, \langid{SV}, $4^{th}$ in \langid{MULTI}, \langid{BN}, \langid{ES}, $5^{th}$ in \langid{PT}, \langid{FA}, \langid{FR}, $6^{th}$ in \langid{IT}, $7^{th}$ in \langid{EN}, $8^{th}$ in \langid{ZH}, and $8^{th}$ in \langid{EN}.
They first trained an XLM-RoBERTa model for entity boundary detection, by recognizing entities within the dataset and classifying them using the \texttt{B-ENTITY} and \texttt{I-ENTITY} tags. They employed a pre-trained mGENRE entity linking model to predict the corresponding Wikipedia title and Wikidata ID for each entity span based on its context. Then, they retrieved the \texttt{``part of''}, \texttt{``instance of''}, \texttt{``occupation''} attributes and the article summary from Wikipedia. Finally, they trained a text classification model to categorize each entity boundary into a fine-grained category using the original sentence, entity boundaries and the external knowledge.

\team{Samsung Research China (SRC) - Beijing} \cite{zhang-EtAl:2023:SemEval1} ranked $2^{nd}$ in \langid{EN}.
They fine-tuned a RoBERTa based ensemble system using a variant of dice loss~\cite{li2019dice} to enhance the model's robustness on long tail entities. In their case dice loss uses soft probabilities over classes, to avoid the model overfitting on the more frequent classes.
Additionally, a Wikipedia knowledge retrieval module was built to augment the sentences with Wikipedia passages.

\team{Sakura}~\cite{poncelas-tkachenko-htun:2023:SemEval} ranked $5^{th}$ in \langid{ES}, $6^{th}$ in \langid{BN}, \langid{DE}, \langid{HI}, \langid{UK}, $7^{th}$ in \langid{IT}, \langid{SV}, \langid{MULTI}, $8^{th}$ in \langid{FA}, $9^{th}$ in \langid{PT}, \langid{ZH}, and $11^{th}$ in \langid{EN}.
They used mBART-50 \cite{https://doi.org/10.48550/arxiv.2008.00401} to translate data from a source language to other target languages part of the shared task. Then, they aligned the tokens using SimAlign \cite{jalili-sabet-etal-2020-simalign} to annotate the entity tokens in the target language. Using the translated examples they increased the training data size between 30K to 102K sentences depending on the language, providing them with a 1\% increase in terms of macro-F1.

\team{KDDIE}~\cite{martin-yang-hsu:2023:SemEval} ranked $5^{th}$ in \langid{EN}.
Using a retrieval index based on Wikipedia they enrich the original training data with additional sentences from Wikipedia. The data is used to train an ensemble of models, and the final NER scores is based on the vote from the different modules such as BERT-CRF, RoBERTa and DeBERTa.

\team{MLlab4CS}~\cite{mukherjee-EtAl:2023:SemEval} ranked $7^{th}$ in \langid{BN}. MuRIL \cite{khanuja2021muril} was fine-tuned with an additional CRF layer used for decoding.
MuRIL is specifically designed to deal with the linguistic characteristics of Indic languages.

\team{CodeNLP}~\cite{marciczuk-walentynowicz:2023:SemEval} ranked $9^{th}$ in \langid{MULTI} and $13^{th}$ in \langid{EN}.
mLUKE-large \cite{yamada-etal-2020-luke} was fine tuned using different data augmentation strategies, where multiple data instances are concatenated as a single input. Their experiments show that the NER model benefits from the additional context, even when the context was unrelated to the original sentence.

\team{silp\_nlp}~\cite{singh-tiwary:2023:SemEval} ranked $7^{th}$ in \langid{HI}, $9^{th}$ in \langid{BN}, $10^{th}$ in \langid{DE, SV}, $11^{th}$ in \langid{ES}, \langid{UK}, $12^{th}$ in \langid{FR}, \langid{IT}, \langid{PT}, $17^{th}$ in \langid{ZH}, $19^{th}$ in \langid{EN}.
Their model is trained in two stages. XLM-RoBERTa is first pre-trained using the multilingual set. Then, the checkpoint is fine-tuned for individual languages.

\team{garNER}~\cite{hossain-EtAl:2023:SemEval} ranked $8^{th}$ in \langid{BN}, $9^{th}$ in \langid{ES}, \langid{SV}, \langid{UK}, \langid{FA}, \langid{HI}, \langid{IT}, $10^{th}$ in \langid{PT}, \langid{FR}, \langid{ZH}, $12^{th}$ in \langid{DE}, \langid{MULTI}, and $16^{th}$ in \langid{EN}.
The authors proposed an approach augmented with external knowledge from Wikipedia. For a given sentence and an entity, the Wikipedia API is called, and the retrieved result is concatenated together with the sentence to provide additional context for token classification. The entities are extracted via spaCy for \langid{English}, and for other languages XLM-RoBERTa is used to detect entities. The authors performed ablation studies to analyze the model performance and found that the relevance of the augmented context is a significant factor in the model's performance. Useful context can help the model to identify some hard entities correctly, while irrelevant context can negatively affect model's predictions.

\team{Sartipi-Sedighin}~\cite{sartipi-EtAl:2023:SemEval} ranked $8^{th}$ in \langid{UK}, $10^{th}$ in \langid{FA}, $11^{th}$ in \langid{BN}, \langid{DE}, \langid{IT}, \langid{PT}, \langid{MULTI}, $12^{th}$ in \langid{ZH}, $13^{th}$ in \langid{HI}, \langid{SV}, $14^{th}$ in \langid{EN}, \langid{ES}, and $16^{th}$ in \langid{FR}.
They used a data augmentation approach, where for entities in the training dataset, additional sentences from  Wikipedia are retrieved. The retrieved sentences are used as additional context. Then, they experimented with Transformer based model variations fine-tuned on different languages. Data augmentation helped their model in certain classes, but negatively impacted some other classes by increasing false negatives, e.g. \symptom.

\team{MaChAmp}~\cite{vandergoot:2023:SemEval} ranked $8^{th}$ in \langid{MULTI}.
mLUKE-large\cite{yamada-etal-2020-luke} was fine-tuned on data coming from all SemEval2023 text based tasks. For NER a CRF decoding layer used. For hyper-parameters they relied on the MaChAmp toolkit \cite{van-der-goot-etal-2021-massive}. They also experimented with separate decoders for each language, using intermediate task pre-training with other SemEval tasks, but did not find it useful for further improvements.

\team{D2KLab}~\cite{ehrhart-plu-troncy:2023:SemEval} ranked $9^{th}$ in \langid{DE}, $10^{th}$ in \langid{ES}, \langid{IT}, \langid{UK}, $11^{th}$ in \langid{FA}, \langid{FR}, $12^{th}$ in \langid{HI, SV}, $13^{th}$ in \langid{PT}, $14^{th}$ in \langid{BN}, \langid{MULTI}, $16^{th}$ in \langid{ZH},  and $18^{th}$ in \langid{EN}.
T-NER library \cite{ushio-camacho-collados-2021-ner} was used to fine-tune a Transformer model. They additionally used 10 other publicly available NER datasets, in addition to the data from MultiCoNER 2 and MultiCoNER.

\team{ERTIM}~\cite{deturck-EtAl:2023:SemEval} ranked $9^{th}$ in \langid{FR}, $11^{th}$ in \langid{ZH}, $12^{th}$ in \langid{FA}, and $20^{th}$ in \langid{EN}.
They fine-tuned different models for the different languages, e.g. BERT, DistilBERT, CamemBERT, and XLM-RoBERTa. Additionally, each input sentence is enriched with relevant Wikipedia articles  for additional context. Furthermore, they annotated a set of additional Farsi sentences extracted from news articles, which provides their system with an improvement of 4.2\% in terms of macro-F1 for \langid{FA}.

\team{LSJSP}~\cite{chatterjee-EtAl:2023:SemEval} ranked $11^{th}$ in \langid{SV}, $13^{th}$ in \langid{ES, UK}, $14^{th}$ in \langid{FR}, $15^{th}$ in \langid{HI, PT}, $16^{th}$ in \langid{BN, MULTI}, and $22^{nd}$ in \langid{EN}.
They rely on a nearest neighbor search method, based on FastText’s \cite{bojanowski2016enriching} implementation, to deal with noisy entities in the dataset.
Next, they use pre-trained transformer models, with a CRF layer for NER prediction.

\team{LLM-RM}~\cite{mehta-varma:2023:SemEval} ranked $11^{th}$ in \langid{HI}, $14^{th}$ in \langid{DE}, $15^{th}$ in \langid{ES}, $27^{th}$ in \langid{EN} by fine-tuning XLM-RoBERTa.

\team{MEERQAT-IRIT}~\cite{lovonmelgarejo-EtAl:2023:SemEval} ranked $12^{th}$ in \langid{ES}, $13^{th}$ in \langid{FR}, $14^{th}$ in \langid{PT}, $21^{st}$ in \langid{EN}.
First, they developed hand-crafted tag descriptors for the fine-grained classes, then, an ensemble representation using the original input and the tag descriptors are used as input to the final CRF layer on top of XLM-RoBERTa.

\team{RIGA}~\cite{mukans-barzdins:2023:SemEval} ranked $12^{th}$ in \langid{EN}.
The original data was augmented using GPT-3 to obtain additional context information, then XLM-RoBERTa (large) was fine-tuned using the adapter fusion approach \cite{pfeiffer-etal-2021-adapterfusion}. The additional context extracted through GPT-3 provides them with a performance boost of 4\% in terms of macro-F1. The context is separated from the input sentence using the separator token \texttt{[SEP]}.

\team{VBD\_NLP}~\cite{hoang-thanh-trieu:2023:SemEval} ranked  $12^{th}$ in \langid{BN} and $26^{th}$ in \langid{EN}.
First, training data was augmented based on BabelNet and Wikipedia redirects to automatically annotate named entities from Wikipedia articles. Then, mDeBERTaV3 with a BiLSTM-CRF layer was fine-tuned for NER. While their model outperformed the baseline in Bangla, it underperformed in English. %

\team{SAB}~\cite{biales:2023:SemEval} ranked $29^{th}$ in \langid{EN}, $17^{th}$ in \langid{ES}, $14^{th}$ in \langid{ES}, \langid{HI}, $12^{th}$ in \langid{UK}, $16^{th}$ in \langid{PT}, $15^{th}$ in \langid{FR}, \langid{DE}, \langid{BN}, \langid{MULTI}, $13^{th}$ in \langid{FA}, \langid{IT}, $18^{th}$ in \langid{ZH}. First, POS tags and dependency relation tags are obtained from open-sourced tools for all languages except \langid{BN} and \langid{MULTI} track.  XLM-R (base) was fine-tuned under a multi-task setup where POS tags, dependency relations and NER labels are predicted. However, they found that using POS and dependency relation did not improve the results.

\team{FII\_Better}~\cite{lupancu-EtAl:2023:SemEval} ranked $13^{th}$ in \langid{DE}, $14^{th}$ in \langid{IT}, $15^{th}$ in \langid{SV}, $16^{th}$ in \langid{ES}, and $17^{th}$ in \langid{EN}.
A BERT model was fine-tuned to label each input token for NER.

\team{IXA}~\cite{andressantamaria:2023:SemEval} ranked  $14^{th}$ in \langid{FA}, \langid{UK}, $15^{th}$ in \langid{IT}, $16^{th}$ in \langid{SV}, $17^{th}$ in \langid{DE}, \langid{HI}, \langid{FR}, \langid{PT}, $18^{th}$ in \langid{BN}, \langid{ES}, $22^{nd}$ in \langid{ZH}, and $34^{th}$ in \langid{EN}. XLM-RoBERTa was fine-tuned for each track separately.

\team{Janko}~\cite{li-guan-ding:2023:SemEval} ranked $14^{th}$ in \langid{ZH}. 
The authors use the last layer of BERT embeddings to represent input tokens, which is then used in a Bi-LSTM model for NER. Additionally, a dropout layer is added, namely R-DROP.

\team{IITD}~\cite{choudhary-chatterjee-saha:2023:SemEval} ranked $15^{th}$ in \langid{EN}.
A two-stage pipeline to fine-tune BERT is proposed: the model is first trained with focal loss to avoid class imbalance issues~\cite{DBLP:journals/corr/abs-1708-02002}. Then, each input is augmented with sentences retrieved from MS-MARCO~\cite{nguyen2016ms} and KILT~\cite{petroni2020kilt} datasets.

\team{YUNLP}~\cite{li:2023:SemEval} ranked $15^{th}$ in \langid{ZH}.
A BERT based approach with a top CRF layer for the NER tag prediction was used. Additionally, a R-Drop layer for regularization to increase the model's robustness was used.

\team{L3i++}~\cite{gonzalezgallardo-EtAl:2023:SemEval} ranked $16^{th}$ in \langid{DE}, \langid{ES}, \langid{HI}, $17^{th}$ in \langid{BN}, $18^{th}$ in \langid{FA}, \langid{FR}, \langid{MULTI}, $20^{th}$ in \langid{IT}, \langid{PT}, \langid{UK}, \langid{SV}, \langid{ZH},  and $25^{th}$ in \langid{EN}.
They submitted three systems. The first model is built with stacked Transformer blocks on top of the BERT encoder with an additional conditional CRF layer. The second one approached the problem with a seq2seq framework: sentences and statement templates filled by candidate named entity span are regarded as the source sequence and the target sequence. In the third approach they transformed NER into a QA task, where a prompt is generated for each type of named entity. The third approach showed strong performance in recall but overall performance was better using the stacked approach.

\team{RGAT}~\cite{chakraborty:2023:SemEval} ranked $23^{rd}$ in \langid{EN}.
They used dependency parse trees from sentences and encode them using a graph attention network. The node representations were computed by taking into account the neighboring nodes and the dependency type. Additionally, they used features from BERT to make the final prediction for a token.

\team{CLaC}~\cite{verma-bergler:2023:SemEval} ranked $24^{th}$ in \langid{EN}. They fine-tuned XLM-RoBERTa, finding that the span prediction approach is better than the sequence labeling approach.

\team{Minanto}~\cite{hfer-mottahedin:2023:SemEval} ranked $28^{th}$ in \langid{EN}.
XLM-RoBERTa was trained using the training data and a set of translated data from CoNLL 2003 and WNUT 2016 datasets.

\section{Insights from the Systems}\label{sec:results}

\paragraph{Integrating External Knowledge:} To overcome the challenges of complex entities, unseen entities, and low context, the integration of external data was a common theme among the submitted systems, similar to the prior edition.
However, this time we observed many new and diverse knowledge sources and novel ways to inject the data into the models for NER prediction.
For example, apart from using paragraphs retrieved from Wikipedia using search engine, participating teams used Wikidata, Wikipedia Infoboxes, and ConceptNet.
Some of these approaches used knowledge sources to compute better representation of the entity labels.
\vspace{-5pt}
\paragraph{Multilingual Models:} Most participants in the multilingual track opted to use the task's baseline model, XLM-RoBERTa. Additionally, some participants used mLUKE, mDEBERTA, and mBERT. In terms of external multilingual resources, participants made mostly use of Wikipedia. 

\vspace{-3pt}
\paragraph{Complex Entities:} Our task includes several classes with complex entities such as media titles. The most challenging entities at the coarse level were from \classname{PROD} class, where the average macro-F1 score across all participants was 0.68. This classes contains challenging entities, with highly complex and ambiguous surface forms, such as \clothing, where the average across all participants was macro-F1=0.58.
There is a high variation among on the challenging coarse types, such as \classname{PROD}. For instance, for \langid{EN} the top ranked system, DAMO-NLP, achieves an F1 of 0.88, while the lowest ranking system IXA achieves a F1 of 0.21. This is highly related to whether the systems used external knowledge.

\begin{figure}[ht!]
    \centering
    \includegraphics[width=.65\columnwidth]{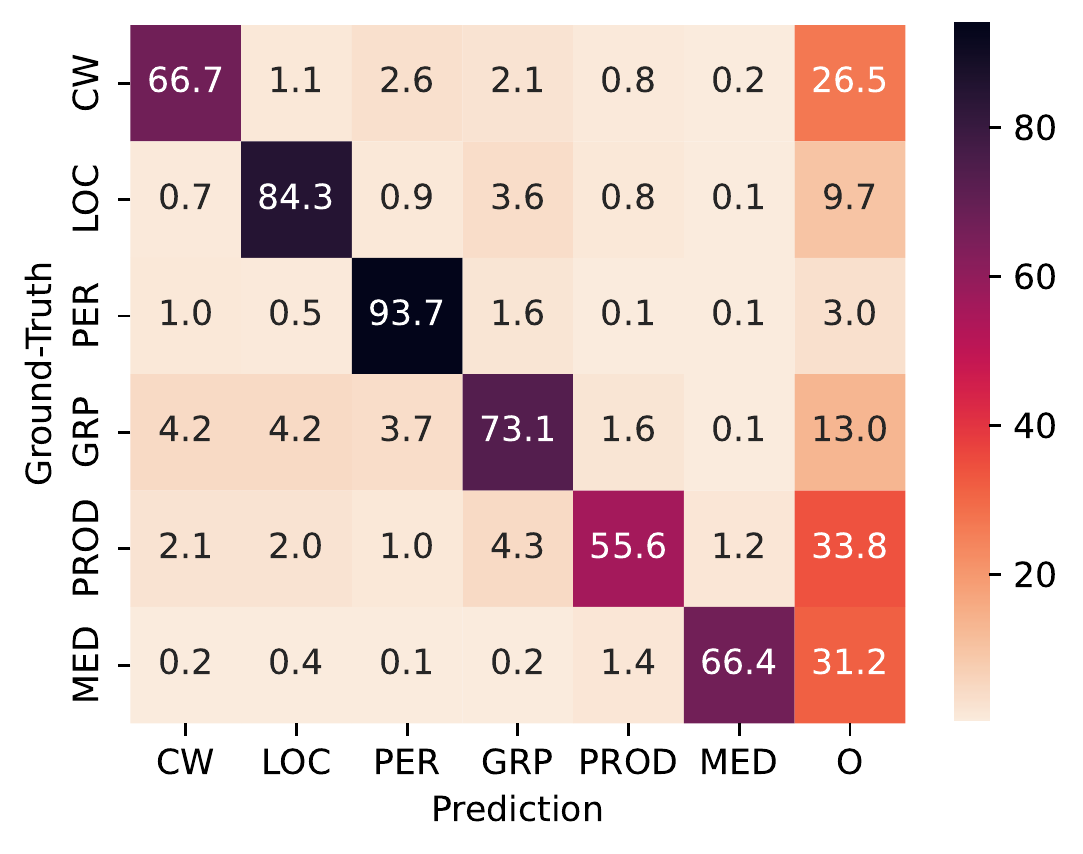}
    \vspace{-15pt}
    \caption{\small Confusion matrix of baseline performance computed at the coarse type level for the \langid{EN} test set.}
    \label{fig:coarse_baseline}
\end{figure}

\Cref{fig:coarse_baseline} shows a confusion matrix of coarse-grained performance. We note that \classname{PROD}, \classname{MED} and \classname{CW} have low recall with more than 25\% of the entities not being identified correctly. \classname{GRP} is misclassified in 4.2\% of the cases with other types such as \classname{LOC} or \classname{CW}, highlighting the surface form ambiguity of this type. On the other hand, \classname{PER} obtains the highest score with 93.7\%, yet at fine-grained level often there is confusion among the different \classname{PER} fine-grained types.

\vspace{-5pt}
\paragraph{Impact of Fine-grained classes:} For coarse types  such as \classname{PER}, participants obtain very high scores, e.g. DAMO-NLP obtains an F1 of 0.97 on the noise-free test set. However, if we inspect the performance at the fine-grained level we notice high variance. For instance, \scientist and \otherper obtain significantly lower scores with F1 scores of 0.70. This gap provides two main insights. First, while the \classname{PER} class is often very easy to spot, distinguishing the more fine-grained types is much more challenging given their high \emph{ambiguity}. Second, for fine-grained NER, capturing context is important. In this case we see that for a class like \scientist, where its entities are often in scientific reporting context (e.g. research breakthroughs), pre-trained Transformer models often confuse such entities as either \artist or \politician, for which such models have much more pretrained knowledge.
\Cref{sec:fine_grained_error_analysis} provides an in-depth error analysis at the fine-grained entity type level for all coarse grained types.

\vspace{-3pt}
\paragraph{Impact of Noise:} 
Evaluation on the noisy subsets shows that most of the participants were impacted significantly.  Comparing the difference in terms of macro-F1 on the noisy and the clean subsets, we notice that across all participants and languages, there is an average performance drop of 10\%. The most impact is observed for \langid{ZH}, where the gap can be as high as macro-F1 = $\blacktriangledown$ 48\%.

Finally, we note that noise is mostly harmful when it affects named entity tokens, while noise on other has a minor impact in terms of NER performance. Across all participants and languages, the average performance dropped 11.1\% when corruption was applied to entity tokens and 4.3\% when it was applied to context tokens.

{

\linepenalty=10000
\vspace{-3pt}
\looseness=-2
\paragraph{ChatGPT and LLMs:} \hspace{-0.2cm} Our evaluation concluded in Jan 2023, and participants did not use ChatGPT for the submissions. \team{DAMO-NLP}~\cite{tan-EtAl:2023:SemEval} reported that the performance of ChatGPT on \langid{MULTI} track is poor and it only achieved 14.78\% F1 score. This matches the results of \citet{lai2023chatgpt} where they evaluated ChatGPT on MultiCoNER task from last year~\cite{multiconer-report}.
\looseness=-2
}

\section{Conclusion}
We presented an overview of the SemEval shared task on identifying complex entities in multiple languages.
We received system submissions from 47 teams, and 34 system papers.
On average, the wining systems for all tracks outperformed the baseline system by a large margin of 35\% F1.

All top-performing teams in \mconer utilized external knowledge bases like Wikipedia and gazetteers to provide additional context. 
We have also observed systems that provided information about the entity classes to help models know the definition of the entity.
In terms of modeling, ensemble strategies helped the systems achieve strong performance. Finally, the impact of noise was significant for all submitted systems, with the macro-F1 dropping significantly when compared between the noisy and clean subsets of test data.
\bibliography{anthology,custom,2023.semeval-1.0-fixed.bib}
\bibliographystyle{acl_natbib}
\clearpage
\newpage
\appendix

\onecolumn
\section*{Appendix}

\section{Detailed Results for Noisy Test Sets}
\label{sec:app_detail_results}

In this section, we provide the detailed performance for a subset of the monolingual tracks that contain a noisy test subset.
For each team, we report the F1 scores for the clean subset and the subset with entity level and context level noise.
\begin{itemize}
 \setlength\itemsep{0em}
    \item Table \ref{tab:full_rank_en} English (EN)
    \item Table \ref{tab:full_rank_it} Italian (IT)
    \item Table \ref{tab:full_rank_es} Spanish (ES)
    \item Table \ref{tab:full_rank_fr} French (FR)
    \item Table \ref{tab:full_rank_pt} Portuguese (PT)
    \item Table \ref{tab:full_rank_sv} Swedish (SV)
    \item Table \ref{tab:full_rank_zh} Chinese (ZH)
\end{itemize}

\begin{table*}[!hbtp]
\resizebox{\textwidth}{!}{%
\begin{tabular}{c lllllll}

\toprule

\textbf{Rank} & 
\textbf{Team} & 
\textbf{Clean Subset F1} & 
\textbf{Noisy Subset F1} & 
\textbf{Entity Noise F1} &
\textbf{Context Noise F1} &  
\textbf{Macro F1}\\ 
\addlinespace[0.2cm]
\midrule

1 & DAMO-NLP & 88.13 & 79.76 & 79.07 & 86.30 & 85.53 \\
2 & {\small Samsung Research China - Beijing} & 85.36 & 77.94 & 77.33 & 83.74 & 83.09 \\
3 & PAI & 86.16 & 65.41 & 63.23 & 84.74 & 80.00 \\
4 & CAIR-NLP & 81.29 & 74.89 & 74.58 & 77.83 & 79.33 \\
5 & KDDIE & 80.08 & 73.50 & 73.03 & 78.05 & 78.06 \\
6 & SRCB & 79.74 & 66.21 & 65.06 & 76.56 & 75.62 \\
7 & IXA/Cogcomp & 76.64 & 64.36 & 63.81 & 69.59 & 72.82 \\
8 & USTC-NELSLIP & 74.87 & 65.76 & 65.35 & 69.18 & 72.15 \\
9 & NLPeople & 76.00 & 62.23 & 60.93 & 74.52 & 71.81 \\
10 & BizNER & 72.12 & 66.64 & 66.32 & 69.65 & 70.44 \\
11 & Sakura & 72.86 & 64.06 & 63.79 & 66.39 & 70.16 \\
12 & RIGA & 70.74 & 66.07 & 65.84 & 68.23 & 69.30 \\
13 & CodeNLP & 66.04 & 57.84 & 57.58 & 60.17 & 63.51 \\
14 & Sartipi-Sedighin & 67.10 & 54.56 & 53.68 & 62.86 & 63.25 \\
15 & IITD & 67.52 & 53.59 & 52.82 & 60.47 & 63.21 \\
16 & garNER & 65.25 & 56.96 & 56.73 & 58.90 & 62.73 \\
17 & FII\_Better & 65.67 & 52.74 & 51.87 & 60.60 & 61.75 \\
18 & D2KLab & 64.72 & 53.54 & 53.12 & 57.24 & 61.29 \\
19 & silp\_nlp & 62.59 & 56.96 & 56.91 & 57.22 & 60.85 \\
20 & Ertim & 61.85 & 52.78 & 52.76 & 52.75 & 59.03 \\
21 & MEERQAT-IRIT & 60.46 & 54.72 & 54.68 & 55.03 & 58.70 \\
22 & LSJSP & 60.67 & 50.48 & 50.44 & 50.59 & 57.51 \\
23 & RGAT & 61.29 & 47.15 & 46.56 & 52.04 & 56.91 \\
24 & CLaC & 57.68 & 49.06 & 48.91 & 50.26 & 55.05 \\
25 & L3i++ & 55.87 & 46.70 & 46.47 & 48.68 & 53.00 \\
26 & VBD\_NLP & 57.00 & 42.44 & 41.45 & 51.06 & 52.65 \\
27 & LLM-RM & 54.73 & 46.30 & 46.17 & 47.45 & 52.08 \\
28 & Minanto & 53.43 & 47.00 & 47.03 & 46.48 & 51.47 \\
29 & SAB & 54.28 & 44.96 & 44.82 & 46.16 & 51.41 \\
30 & ShathaTaymaaTeam & 52.34 & 44.78 & 45.09 & 41.67 & 50.02 \\
31 & azaad@BND & 50.09 & 41.28 & 41.09 & 42.94 & 47.42 \\
32 & LISAC FSDM-USMBA & 47.36 & 36.58 & 36.27 & 39.32 & 44.00 \\
33 & YNU-HPCC & 29.95 & 25.31 & 25.31 & 25.21 & 28.52 \\
34 & IXA & 16.88 & 11.84 & 11.49 & 15.09 & 15.39 \\
\bottomrule

\end{tabular}%
} 
\caption{Detailed results for the English track.}
\label{tab:full_rank_en}
\end{table*}

\begin{table*}[!hbtp]
\resizebox{\textwidth}{!}{%
\begin{tabular}{c lllllll}

\toprule

\textbf{Rank} & 
\textbf{Team} & 
\textbf{Clean Subset F1} & 
\textbf{Noisy Subset F1} & 
\textbf{Entity Noise F1} &
\textbf{Context Noise F1} &  
\textbf{Macro F1}\\ 
\addlinespace[0.2cm]
\midrule

1 & DAMO-NLP & 91.85 & 85.89 & 85.30 & 90.99 & 89.79 \\
2 & PAI & 88.94 & 76.53 & 75.14 & 89.56 & 84.88 \\
3 & CAIR-NLP & 85.08 & 81.00 & 80.58 & 84.63 & 83.78 \\
4 & BizNER & 77.24 & 74.81 & 74.34 & 79.33 & 76.48 \\
5 & USTC-NELSLIP & 78.06 & 70.65 & 70.05 & 75.91 & 75.70 \\
6 & IXA/Cogcomp & 78.16 & 67.66 & 66.77 & 75.95 & 74.67 \\
7 & Sakura & 76.67 & 69.03 & 68.53 & 73.18 & 74.19 \\
8 & NLPeople & 77.45 & 65.88 & 64.58 & 78.87 & 73.71 \\
9 & garNER & 70.16 & 63.99 & 63.53 & 67.81 & 68.20 \\
10 & D2KLab & 68.17 & 57.68 & 57.07 & 63.15 & 64.77 \\
11 & Sartipi-Sedighin & 67.61 & 57.95 & 57.16 & 65.31 & 64.50 \\
12 & silp\_nlp & 64.53 & 60.13 & 60.00 & 61.00 & 63.11 \\
13 & SAB & 60.36 & 51.60 & 51.15 & 55.56 & 57.57 \\
14 & FII\_Better & 60.32 & 47.85 & 46.76 & 58.36 & 56.36 \\
15 & IXA & 20.05 & 14.82 & 14.38 & 18.84 & 18.41 \\
\bottomrule

\end{tabular}%
} 
\caption{Detailed results for the Italian track.}
\label{tab:full_rank_it}
\end{table*}

\begin{table*}[!hbtp]
\resizebox{\textwidth}{!}{%
\begin{tabular}{c lllllll}

\toprule

\textbf{Rank} & 
\textbf{Team} & 
\textbf{Clean Subset F1} & 
\textbf{Noisy Subset F1} & 
\textbf{Entity Noise F1} &
\textbf{Context Noise F1} &  
\textbf{Macro F1}\\ 
\addlinespace[0.2cm]
\midrule

1 & DAMO-NLP & 91.74 & 85.81 & 85.29 & 91.06 & 89.78 \\
2 & CAIR-NLP & 85.03 & 80.66 & 80.44 & 82.92 & 83.63 \\
3 & USTC-NELSLIP & 77.25 & 68.52 & 68.00 & 73.73 & 74.44 \\
4 & IXA/Cogcomp & 77.65 & 66.09 & 65.48 & 72.30 & 73.81 \\
5 & Sakura & 75.42 & 67.39 & 66.93 & 72.01 & 72.85 \\
6 & NLPeople & 77.22 & 63.53 & 62.43 & 74.76 & 72.76 \\
7 & PAI & 79.35 & 55.25 & 53.16 & 75.10 & 71.67 \\
8 & BizNER & 72.60 & 69.11 & 69.08 & 69.53 & 71.48 \\
9 & garNER & 66.19 & 58.43 & 58.21 & 60.35 & 63.73 \\
10 & D2KLab & 66.69 & 55.75 & 55.26 & 60.47 & 63.17 \\
11 & silp\_nlp & 64.88 & 58.77 & 58.66 & 59.71 & 62.90 \\
12 & MEERQAT-IRIT & 63.04 & 56.42 & 56.16 & 58.78 & 60.93 \\
13 & LSJSP & 63.39 & 54.46 & 54.26 & 56.21 & 60.55 \\
14 & Sartipi-Sedighin & 62.27 & 50.25 & 49.70 & 55.57 & 58.41 \\
15 & LLM-RM & 57.42 & 49.32 & 49.19 & 50.47 & 54.81 \\
16 & FII\_Better & 58.96 & 44.77 & 43.57 & 56.07 & 54.51 \\
17 & SAB & 50.83 & 42.62 & 42.58 & 42.87 & 48.22 \\
18 & IXA & 17.65 & 12.16 & 11.83 & 14.59 & 16.01 \\
\bottomrule

\end{tabular}%
} 
\caption{Detailed results for the Spanish track.}
\label{tab:full_rank_es}
\end{table*}

\begin{table*}[!hbtp]
\resizebox{\textwidth}{!}{%
\begin{tabular}{c lllllll}

\toprule

\textbf{Rank} & 
\textbf{Team} & 
\textbf{Clean Subset F1} & 
\textbf{Noisy Subset F1} & 
\textbf{Entity Noise F1} &
\textbf{Context Noise F1} &  
\textbf{Macro F1}\\ 
\addlinespace[0.2cm]
\midrule

1 & DAMO-NLP & 91.62 & 85.14 & 84.58 & 90.49 & 89.59 \\
2 & PAI & 89.50 & 78.71 & 77.64 & 88.96 & 86.17 \\
3 & CAIR-NLP & 84.67 & 79.54 & 79.22 & 82.55 & 83.08 \\
4 & BizNER & 79.09 & 75.63 & 75.29 & 78.92 & 78.01 \\
5 & IXA/Cogcomp & 78.60 & 65.81 & 64.93 & 74.14 & 74.52 \\
6 & USTC-NELSLIP & 76.81 & 68.49 & 67.93 & 73.55 & 74.25 \\
7 & Sakura & 75.58 & 66.86 & 66.38 & 71.26 & 72.86 \\
8 & NLPeople & 77.12 & 63.40 & 62.02 & 76.51 & 72.85 \\
9 & Ertim & 69.73 & 58.60 & 57.77 & 66.09 & 66.30 \\
10 & garNER & 68.09 & 60.22 & 59.77 & 64.27 & 65.68 \\
11 & D2KLab & 67.70 & 56.05 & 55.30 & 63.12 & 64.09 \\
12 & silp\_nlp & 64.40 & 58.04 & 57.81 & 60.02 & 62.39 \\
13 & MEERQAT-IRIT & 61.29 & 53.65 & 53.23 & 57.32 & 58.90 \\
14 & LSJSP & 58.74 & 52.60 & 52.34 & 54.93 & 56.83 \\
15 & SAB & 57.98 & 48.61 & 48.19 & 52.41 & 55.07 \\
16 & Sartipi-Sedighin & 56.99 & 50.40 & 50.22 & 52.05 & 54.94 \\
17 & IXA & 18.90 & 13.89 & 13.49 & 17.29 & 17.40 \\
\bottomrule

\end{tabular}%
} 
\caption{Detailed results for the French track.}
\label{tab:full_rank_fr}
\end{table*}

\begin{table*}[!hbtp]
\resizebox{\textwidth}{!}{%
\begin{tabular}{c lllllll}

\toprule

\textbf{Rank} & 
\textbf{Team} & 
\textbf{Clean Subset F1} & 
\textbf{Noisy Subset F1} & 
\textbf{Entity Noise F1} &
\textbf{Context Noise F1} &  
\textbf{Macro F1}\\ 
\addlinespace[0.2cm]
\midrule

1 & DAMO-NLP & 87.33 & 83.38 & 83.04 & 88.00 & 85.97 \\
2 & PAI & 84.56 & 76.12 & 75.50 & 82.87 & 81.61 \\
3 & CAIR-NLP & 81.73 & 77.10 & 76.94 & 78.61 & 80.16 \\
4 & BizNER & 74.36 & 70.35 & 70.12 & 72.81 & 72.97 \\
5 & IXA/Cogcomp & 76.00 & 65.54 & 64.91 & 72.32 & 72.28 \\
6 & USTC-NELSLIP & 74.04 & 65.91 & 65.49 & 70.37 & 71.26 \\
7 & Deep Learning Brasil & 72.07 & 68.91 & 68.75 & 70.11 & 70.97 \\
8 & NLPeople & 74.50 & 62.22 & 61.27 & 73.48 & 70.16 \\
9 & Sakura & 72.74 & 64.76 & 64.29 & 69.57 & 69.98 \\
10 & garNER & 66.81 & 60.04 & 59.82 & 61.52 & 64.51 \\
11 & Sartipi-Sedighin & 63.75 & 56.57 & 56.30 & 59.32 & 61.28 \\
12 & silp\_nlp & 63.07 & 57.23 & 56.99 & 59.93 & 61.05 \\
13 & D2KLab & 64.44 & 53.98 & 53.47 & 59.01 & 60.79 \\
14 & MEERQAT-IRIT & 61.82 & 56.17 & 56.01 & 58.00 & 59.87 \\
15 & LSJSP & 60.63 & 53.60 & 53.32 & 56.23 & 58.23 \\
16 & SAB & 57.55 & 47.56 & 47.21 & 51.08 & 54.12 \\
17 & IXA & 18.40 & 13.91 & 13.61 & 17.95 & 16.97 \\
\bottomrule

\end{tabular}%
} 
\caption{Detailed results for the Portuguese track.}
\label{tab:full_rank_pt}
\end{table*}

\begin{table*}[!hbtp]
\resizebox{\textwidth}{!}{%
\begin{tabular}{c lllllll}

\toprule

\textbf{Rank} & 
\textbf{Team} & 
\textbf{Clean Subset F1} & 
\textbf{Noisy Subset F1} & 
\textbf{Entity Noise F1} &
\textbf{Context Noise F1} &  
\textbf{Macro F1}\\ 
\addlinespace[0.2cm]
\midrule

1 & DAMO-NLP & 91.08 & 86.76 & 86.34 & 91.41 & 89.57 \\
2 & CAIR-NLP & 84.54 & 79.75 & 79.49 & 80.75 & 82.88 \\
3 & IXA/Cogcomp & 80.75 & 68.69 & 67.99 & 74.81 & 76.54 \\
4 & BizNER & 77.23 & 73.87 & 73.54 & 77.78 & 76.12 \\
5 & USTC-NELSLIP & 78.51 & 69.64 & 69.22 & 72.87 & 75.47 \\
6 & NLPeople & 79.31 & 67.15 & 66.30 & 75.22 & 75.08 \\
7 & Sakura & 76.74 & 68.12 & 67.66 & 71.74 & 73.79 \\
8 & PAI & 81.53 & 55.22 & 53.04 & 77.04 & 72.38 \\
9 & garNER & 70.40 & 62.19 & 61.86 & 66.01 & 67.63 \\
10 & silp\_nlp & 67.15 & 60.87 & 60.53 & 63.74 & 65.00 \\
11 & LSJSP & 67.23 & 58.63 & 58.18 & 64.13 & 64.36 \\
12 & D2KLab & 66.78 & 55.80 & 55.29 & 61.14 & 62.98 \\
13 & Sartipi-Sedighin & 64.69 & 56.57 & 56.15 & 60.38 & 61.95 \\
14 & SAB & 61.58 & 51.14 & 50.90 & 52.90 & 58.03 \\
15 & FII\_Better & 56.66 & 43.11 & 42.20 & 50.67 & 52.12 \\
16 & IXA & 27.96 & 21.72 & 21.30 & 26.72 & 25.96 \\
\bottomrule

\end{tabular}%
} 
\caption{Detailed results for the Swedish track.}
\label{tab:full_rank_sv}
\end{table*}

\begin{table*}[!hbtp]
\resizebox{\textwidth}{!}{%
\begin{tabular}{c lllllll}

\toprule

\textbf{Rank} & 
\textbf{Team} & 
\textbf{Clean Subset F1} & 
\textbf{Noisy Subset F1} & 
\textbf{Entity Noise F1} &
\textbf{Context Noise F1} &  
\textbf{Macro F1}\\ 
\addlinespace[0.2cm]
\midrule

1 & NetEase.AI & 88.47 & 69.05 & 67.80 & 86.43 & 84.05 \\
2 & DAMO-NLP & 82.91 & 54.32 & 52.26 & 81.45 & 75.98 \\
3 & SRCB & 87.19 & 39.39 & 35.37 & 88.37 & 75.86 \\
4 & PAI & 86.23 & 41.90 & 38.57 & 85.31 & 74.87 \\
5 & Taiji & 75.70 & 61.39 & 60.53 & 72.28 & 72.52 \\
6 & USTC-NELSLIP & 70.10 & 55.06 & 54.16 & 64.75 & 66.57 \\
7 & NLPeople & 71.43 & 48.95 & 47.91 & 61.57 & 65.96 \\
8 & IXA/Cogcomp & 70.35 & 48.37 & 47.36 & 59.88 & 64.86 \\
9 & Sakura & 68.79 & 51.20 & 50.43 & 59.73 & 64.61 \\
10 & garNER & 67.50 & 50.17 & 49.57 & 58.23 & 63.47 \\
11 & Ertim & 64.26 & 44.38 & 43.26 & 59.44 & 59.45 \\
12 & Sartipi-Sedighin & 62.60 & 46.46 & 46.10 & 49.10 & 58.70 \\
13 & CAIR-NLP & 62.89 & 44.74 & 43.84 & 56.16 & 58.43 \\
14 & Janko & 62.45 & 44.70 & 44.14 & 52.15 & 57.90 \\
15 & YNUNLP & 61.45 & 42.69 & 42.17 & 50.29 & 56.57 \\
16 & D2KLab & 58.75 & 43.34 & 42.67 & 48.12 & 54.92 \\
17 & silp\_nlp & 54.65 & 42.11 & 41.57 & 48.95 & 51.65 \\
18 & SAB & 47.71 & 33.37 & 32.46 & 42.82 & 44.12 \\
19 & NCUEE-NLP & 51.36 & 18.24 & 15.39 & 43.74 & 44.09 \\
20 & L3i++ & 38.02 & 27.13 & 26.63 & 32.99 & 35.34 \\
21 & YNU-HPCC & 34.24 & 24.07 & 23.52 & 32.50 & 31.66 \\
22 & IXA & 8.06 & 4.49 & 4.35 & 5.20 & 6.93 \\
\bottomrule

\end{tabular}%
} 
\caption{Detailed results for the Chinese track.}
\label{tab:full_rank_zh}
\end{table*}

\newpage
\section{Fine-Grained Results Analysis}\label{sec:fine_grained_error_analysis}

\begin{figure*}[ht!]
    \centering
    \includegraphics[width=\columnwidth]{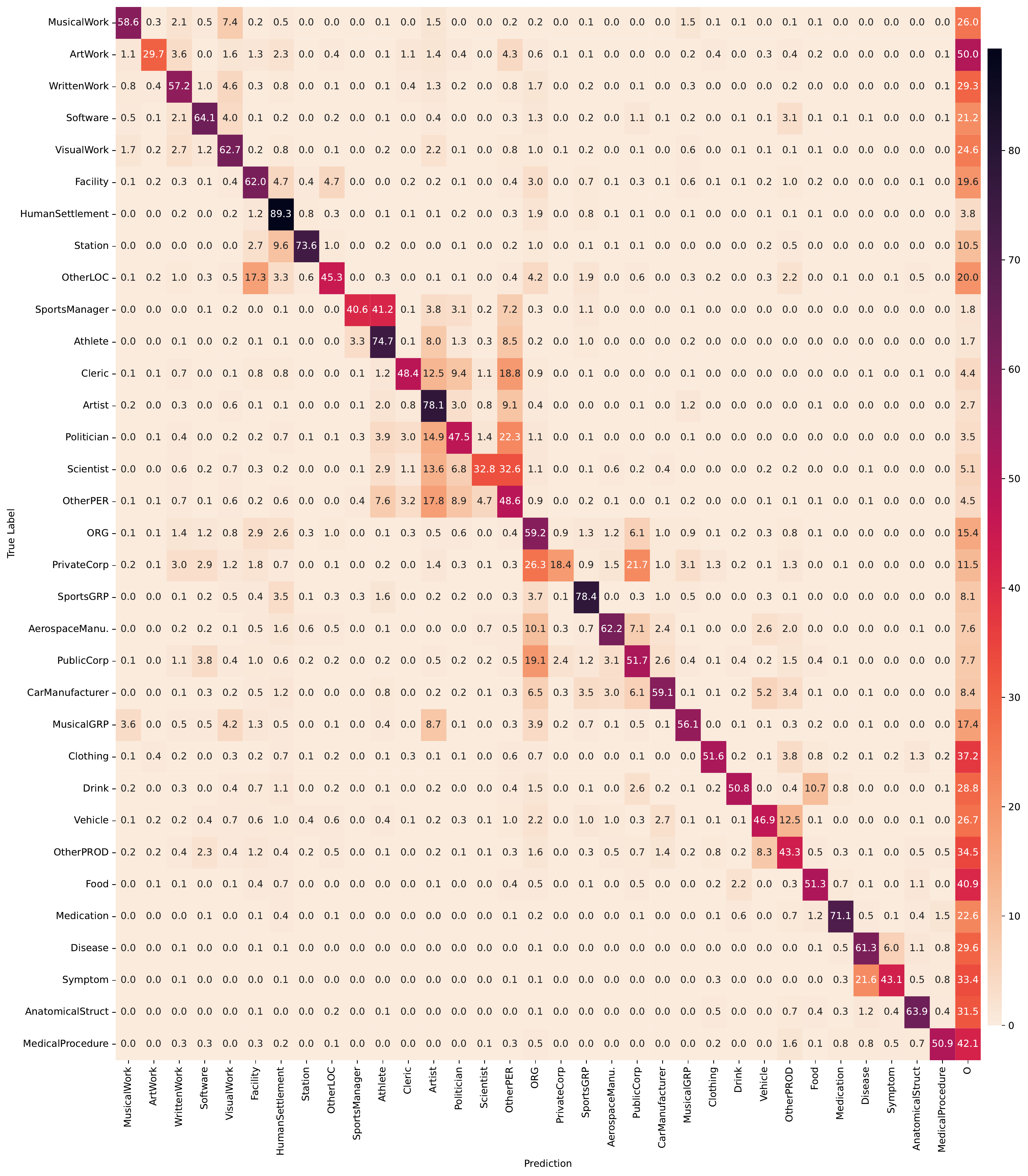}
    \caption{Confusion matrix of baseline performance computed at the fine-grained level for the \langid{EN} test set.}
    \label{fig:baseline_all_classes}
\end{figure*}

Figure~\ref{fig:baseline_all_classes} shows the misclassification across the different fine-grained types for the baseline approach on the \langid{EN} test set. An ideal classifier would have a 100\% performance on the diagonal.

\paragraph{\classname{CW}.} For this class, the baseline has low recall, with many of the entities being missed (\classname{O} tag). In terms of misclassifying the fine-grained types, we note that the highest confusion is between \musicalwork and \visualwork, with 7.4\% of false positives. %

\paragraph{\classname{GRP}.} In the case of \classname{GRP}, we notice a high confusion between \org, \publiccorp and \privatecorp, with error rates going up to 26.3\%. This highlights the difficulty of the different fine-grained classes, where context capture is important. 
Even more importantly in this particular problem of fine-grained NER, external knowledge or world knowledge of entities is crucial to distinguish between such fine-grained differences.
In this case, external knowledge about different corporations may be necessary to correctly distinguish between different named entity types.

\paragraph{\classname{LOC}.} For this class, most of the errors are between \facility and \otherloc. 

\paragraph{\classname{PER}.} In the case of \classname{PER}, \sportsmanager is confused as \athlete in 41.2\% of the cases (this is because many sports managers are former athletes). The \classname{PER} coarse type is highly challenging in some of the fine-grained types, given that the surface forms can be highly ambiguous, and only the context can differentiate between the different types (\athlete, \scientist, \artist, etc.)

\paragraph{\classname{MED}.} In this case, we notice a high confusion between \disease and \symptom, with 21.6\%. This is an interesting insights, given that often, names for diseases and symptoms are used interchangeably (i.e., a symptom may cause a disease that is referred using the same name). 

\paragraph{\classname{PROD}.} Finally, here we notice that \drink and \food are often confused with each other with 10.7\%. This highlights some of the ambiguous cases where a drink may be considered both, e.g. \emph{milk}. Finally, the most misclassification happen between \vehicle and \otherprod. A potential cause for this is the lack of detailed type assignment of entities in Wikidata, which may lead to such misclassifications, i.e. \otherprod entities may actually belong to \vehicle, however they are not explicitly associated with this type in Wikidata.

\end{document}